%% file: main.tex
\documentclass[10pt,twocolumn,letterpaper]{article}

\usepackage[pagenumbers]{cvpr} 

\input{preamble}

\definecolor{cvprblue}{rgb}{0.21,0.49,0.74}
\usepackage[pagebackref,breaklinks,colorlinks,citecolor=cvprblue]{hyperref}

\usepackage{multirow}
\usepackage{bm}
\usepackage{bbm}

\usepackage{times}
\usepackage{epsfig}
\usepackage{graphicx}
\usepackage{amsmath}
\usepackage{amssymb}
\usepackage{algorithm}
\usepackage{algorithmic}

\def\paperID{8847} 
\def\confName{CVPR}
\def\confYear{2024}

\title{Dual Relation Mining Network for Zero-Shot Learning}

\author{Jinwei Han$^{1}$, Yingguo Gao$^2$, Zhiwen Lin$^2$, Ke Yan$^2$, Shouhong Ding$^2$, Yuan Gao$^1$, Gui-Song Xia$^{1\dag}$\\
$^1$Wuhan University $^2$YouTu Lab, Tencent\\
{\tt\small \{hanjinwei, guisong.xia\}@whu.edu.cn, \{ethan.y.gao\}@gmail.com}\\
{\tt\small xavier.lin@foxmail.com, \{yingguogao, kerwinyan, ericshding\}@tencent.com}\\
}

\begin{document}
\maketitle

\newcommand\blfootnote[1]{%
\begingroup
\renewcommand\thefootnote{}\footnote{#1}%
\addtocounter{footnote}{-1}%
\endgroup
}

\blfootnote{$^\dag$ Corresponding author.}

\input{sec/0_abstract}    
\input{sec/1_intro}
\input{sec/2_related_work}
\input{sec/3_method}
\input{sec/4_experiments}
\input{sec/5_conclusion}
{
    \small
    \bibliographystyle{ieeenat_fullname}
    \bibliography{main}
}


\end{document}

%% file: preamble.tex
\usepackage[dvipsnames]{xcolor}


%% file: sec/0_abstract.tex
\begin{abstract}
Zero-shot learning (ZSL) aims to recognize novel classes through transferring shared semantic knowledge (e.g., attributes) from seen classes to unseen classes. Recently, attention-based methods have exhibited significant progress which align visual features and attributes via a spatial attention mechanism. However, these methods only explore visual-semantic relationship in the spatial dimension, which can lead to classification ambiguity when different attributes share similar attention regions, and semantic relationship between attributes is rarely discussed. To alleviate the above problems, we propose a Dual Relation Mining Network (DRMN) to enable more effective visual-semantic interactions and learn semantic relationship among attributes for knowledge transfer. Specifically, we introduce a Dual Attention Block (DAB) for visual-semantic relationship mining, which enriches visual information by multi-level feature fusion and conducts spatial attention for visual to semantic embedding. Moreover, an attribute-guided channel attention is utilized to decouple entangled semantic features. For semantic relationship modeling, we utilize a Semantic Interaction Transformer (SIT) to enhance the generalization of attribute representations among images. Additionally, a global classification branch is introduced as a complement to human-defined semantic attributes, and we then combine the results with attribute-based classification. Extensive experiments demonstrate that the proposed DRMN leads to new state-of-the-art performances on three standard ZSL benchmarks, i.e., CUB, SUN, and AwA2.
\end{abstract}

%% file: sec/1_intro.tex
\section{Introduction}

\begin{figure}[t]
\begin{center}
\resizebox{1\linewidth}{!}{
\includegraphics[trim={300pt 35pt 300pt 35pt},clip,width=1\linewidth]{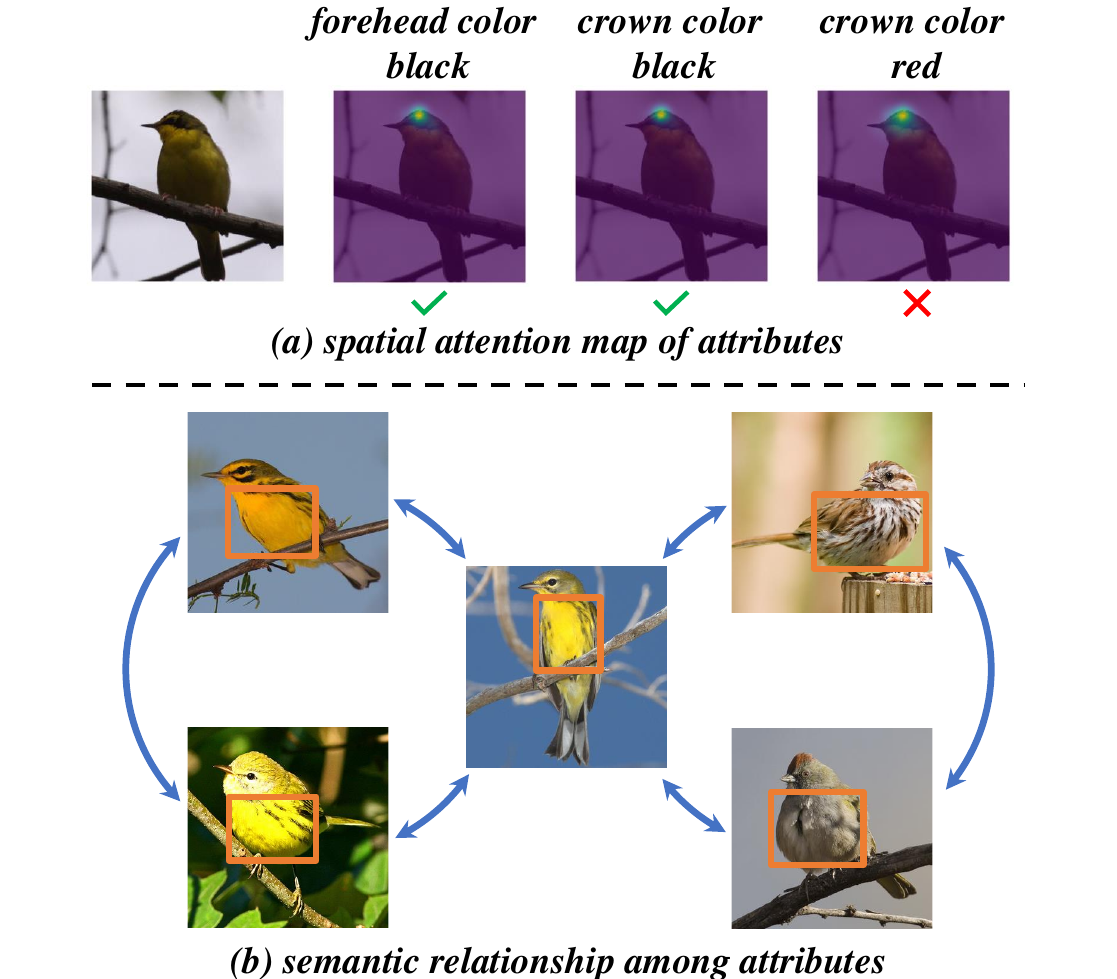}
}
\end{center}
\caption{Motivation illustration. (a) Different attributes may share similar attention areas, posing challenges for attribute prediction. (b) The appearance of the same attribute can vary, while different attributes may share similar semantic information that can be leveraged to facilitate knowledge transfer.}
\label{fig:intro}
\end{figure}

Visual recognition techniques \cite{he2016deep,dosovitskiy2020image} have shown remarkable achievements in identifying the classes they are trained on, while their generalization to novel classes or concepts is still insufficient. To tackle this challenge, zero-shot learning (ZSL) has been proposed and aims to recognize new classes through leveraging shared semantic descriptions (\textit{e.g.}, attributes) between seen and unseen classes~\cite{lampert2009learning, palatucci2009zero}. Depending on the availability of seen classes during the test phase, ZSL can be categorized into Conventional ZSL (CZSL) and Generalized ZSL (GZSL). In contrast to CZSL which only predicts unseen classes, GZSL is more practical where test images belong to both seen and unseen classes.

One of the main challenges in ZSL is to establish an effective mapping between visual and semantic spaces on seen classes, thereby facilitating knowledge transfer to unseen classes. In this regard, numerous ZSL approaches have been proposed. Generative-based methods predominantly rely on generative adversarial networks (GANs)~\cite{Xian2018FeatureGN,Li2019LeveragingTI,Yu2020EpisodeBasedPG,Vyas2020LeveragingSA,Chen2021FREE,kong2022compactness} or variational autoencoders (VAEs)~\cite{Arora2018GeneralizedZL,xian2019f,keshari2020generalized,su2022distinguishing} to synthesize unseen features and transform the ZSL problem into a supervised classification task. However, these methods often struggle to synthesize diverse and discriminative features that are consistent with real images. Embedding-based methods~\cite{Zhang2017LearningAD,Li2018DiscriminativeLO,Chen2018ZeroShotVR,xie2019attentive,Liu2019AttributeAF,zhu2019semantic} aim to associate visual features with corresponding semantic features via embedding functions. Among them, attention-based approaches~\cite{Huynh2020FineGrainedGZ,Xu2020AttributePN,Liu2021GoalOrientedGE} have shown great superiority which focus on discriminative image regions and encode these features to achieve effective visual-semantic interaction. They mainly employ spatial attention through attribute localization, but the intricate relationship between visual and semantic features remains under-explored.

First, the distribution of attribute features is multi-level in the visual space, and relying solely on high-level features from the last block of the backbone usually leads to a loss of low-level patterns. Second, although spatial attention helps to localize the most relevant regions for attributes, the extracted features will be not distinguishable enough for classification when different attributes share similar attention regions. As depicted in Fig.\ref{fig:intro}(a), the absent attribute (\textit{i.e.}, crown color red) and the two existing attributes (\textit{i.e.}, forehead color black and crown color black) have a similar attention region in the spatial attention map. Moreover, while most prior methods struggle to exploit the visual-semantic relationship, the semantic relationship among attributes (as shown in Fig.\ref{fig:intro}(b)) is seldom discussed in ZSL. Leveraging semantic relationship facilitates knowledge transfer between data samples~\cite{hou2022batchformer} and can lead to more robust and generalized representations, especially in the presence of data scarcity. 

Based on the aforementioned observations, we propose a Dual Relation Mining Network (DRMN) to jointly mine dual relation (\textit{i.e.}, visual-semantic relationship and semantic-semantic relationship) for ZSL. Specifically, our DRMN comprises a Dual Attention Block (DAB) for visual-semantic relationship mining and a Semantic Interaction Transformer (SIT) for semantic relationship modeling. Within the DAB, we enhance the visual feature through multi-level feature fusion and employ region-attribute spatial attention for visual to semantic embedding. Moreover, an attribute-guided channel attention is utilized to improve the discriminability of attribute features in the channel dimension. To explore semantic relationship, we use SIT to strengthen the generalization of attribute representations among images for more effectively knowledge transfer. Additionally, there may exist latent features which complement the human-defined semantic attributes \cite{peng2017joint,jiang2017learning,Liu2019AttributeAF}, so we introduce a global classification branch and combine the results with attribute-based classification. Extensive experiments demonstrate that our DRMN achieves a new state-of-the-art on three popular ZSL benchmarks, indicating the effectiveness of our insight for dual relation mining.

Our main contributions are summarized as follows: 
\begin{itemize}
    \item We present a novel scheme DRMN for ZSL, which jointly mines visual-semantic and semantic-semantic relationships for effective knowledge transfer.
    \item We propose a Dual Attention Block (DAB) to enrich visual features and conduct visual-semantic interactions via spatial and channel attention. A Semantic Interaction Transformer (SIT) enhances the generalization of attribute representations among images.
    \item We introduce a global classification branch as a complement to human-defined semantic attributes and further combine the results with attribute-based classification to improve the performance.
    \item Extensive experiments demonstrate that our DRMN achieves the new state-of-the-art on three ZSL benchmarks, \ie, CUB \cite{Welinder2010CaltechUCSDB2}, SUN \cite{Patterson2012SUNAD}, and AwA2 \cite{Xian2017ZeroShotLC}.
\end{itemize}

\begin{figure*}
\begin{center}
\resizebox{1\linewidth}{!}{
\includegraphics[trim={150pt 110pt 150pt 110pt},clip,width=1\linewidth]{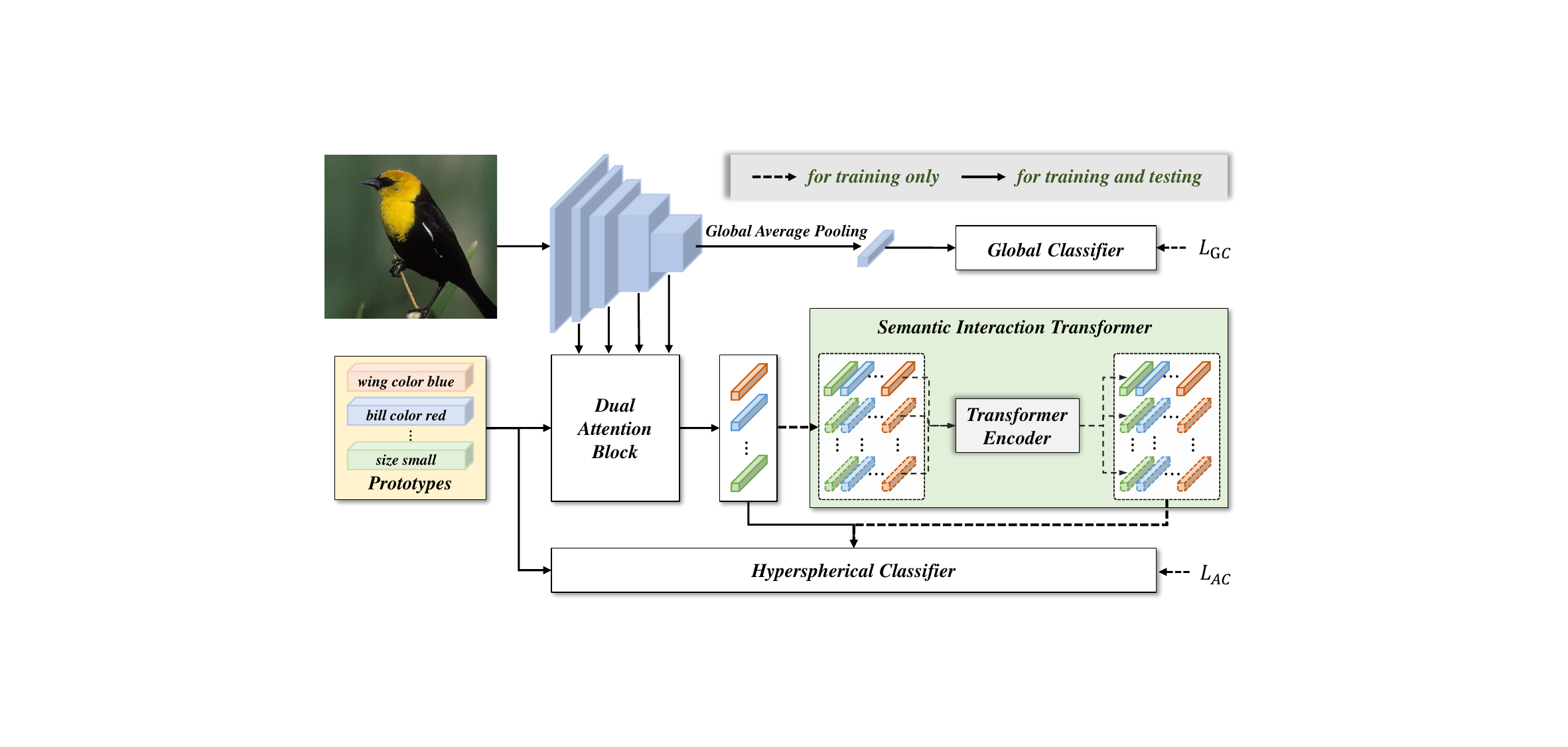}
}
\end{center}
\caption{The architecture of the proposed scheme DRMN. DRMN consists of a Dual Attention Block (DAB), a Semantic Interaction Transformer (SIT), and a global classification branch. DAB fuses multi-level visual features and employs spatial and channel attention mechanisms for visual to semantic embedding. SIT models semantic relationship to enhance the generalization of attribute representations for knowledge transfer. The predicted attributes are projected onto the hyperspherical space for classification. The global classification branch complements the human-defined attributes and we combine the results with attribute-based classification.}
\label{fig:framework}
\end{figure*}

%% file: sec/2_related_work.tex
\section{Related Work}
\label{sec:related work}
\subsection{Zero-Shot Learning}
Zero-shot learning (ZSL) aims to bridge the gap between seen and unseen classes by learning a mapping between visual and semantic space through shared semantic descriptions~\cite{palatucci2009zero,lampert2009learning,pourpanah2022review}. Numerous approaches have been proposed to tackle this challenge, which can be classified into two main categories: generative-based and embedding-based. Generative-based methods mainly employ three types of mainstream generative approaches, \ie, generative adversarial nets (GANs)~\cite{Xian2018FeatureGN,Li2019LeveragingTI,Yu2020EpisodeBasedPG,Vyas2020LeveragingSA,Chen2021FREE,kong2022compactness}, variational autoencoders (VAEs)~\cite{Arora2018GeneralizedZL,xian2019f,keshari2020generalized,su2022distinguishing}, and generative flows \cite{Shen2020InvertibleZR,chen2022gsmflow}. Through synthesizing features of unseen classes based on shared semantic descriptions and features of seen classes, these methods transform the ZSL problem into a traditional classification task. While generative-based methods can alleviate the overfitting to seen classes, they struggle to synthesize diverse and discriminative features consistent with real images, which impairs classification accuracy. In contrast, embedding-based methods~\cite{Zhang2017LearningAD,Chen2018ZeroShotVR,Li2018DiscriminativeLO,Liu2019AttributeAF,zhu2019semantic} are non-generative and learn to associate visual features with corresponding semantic features via embedding functions. However, these methods mainly align global visual features with class semantics, neglecting discriminative regions which are crucial for recognizing fine-grained categories and results in insufficient visual-semantic interactions for knowledge transfer.

\subsection{Attention-based ZSL}
More recently, attention-based methods have shown considerable progress in ZSL by focusing on discriminative regions of images and encoding region features for visual to semantic embedding. Some approaches \cite{Huynh2020FineGrainedGZ,Liu2021GoalOrientedGE,Chen2022TransZeroAT,chen2022msdn} utilize semantic vectors extracted from pretrained language model GloVe \cite{pennington2014glove} to guide the alignment of each region feature with its attribute semantic vector. Others \cite{Xu2020AttributePN,wang2021dual} construct attribute prototypes for direct attribute localization, which improves the locality of image representation and transferability. Meanwhile, RGEN \cite{xie2020region} incorporates the region-based relation reasoning into ZSL to model the relationships among local image regions. Nonetheless, they mainly rely on a spatial attention mechanism between region features and attributes, which can result in challenges for classification when different attributes share similar attention areas. Moreover, semantic relationship between attributes is often not adequately considered. To tackle these challenges, we propose mining visual-semantic relationship to enhance visual to semantic embedding and model semantic-semantic relationship for a more generalized attribute representation for knowledge transfer.

%% file: sec/3_method.tex
\section{Methodology}
\label{sec:method}
\noindent{\bf Overview.}
As shown in Fig.\ref{fig:framework}, our DRMN comprises two branches that share the same backbone for classification. {\em In the attribute-based classification branch}, we propose a Dual Attention Block (DAB) to mine visual-semantic relationships, which enhances visual information through multi-level feature fusion and conducts visual-semantic interactions via spatial and channel attention. To model semantic relationships, we employ a Semantic Interaction Transformer (SIT) to strengthen the consistency of attribute representations for more effective knowledge transfer. The hyperspherical classifier then outputs class logits based on attribute prototypes and the predicted semantic features. {\em In the global classification branch}, we aim to learn latent features as a supplement to the human-defined attributes and integrate the results of the two branches during test phase.

\noindent{\bf Notation.}
Let $\mathcal{D}^{s}=\left\{\left(x_{i}^{s},y_{i}^{s}\right)\right\}$ be a dataset associated with seen classes $\mathcal{C}^s$ , where $x_i^s \in \mathcal{X}^s$ denotes the image $i$ with the class label $y_i^s \in \mathcal{Y}^s$. Similarly, the data for unseen classes $\mathcal{C}^u$ is denoted by $\mathcal{D}^{u}=\left\{\left(x_{i}^{u}, y_{i}^{u}\right)\right\}$, where $x_{i}^{u}\in \mathcal{X}^u$ is an image from one of the unseen classes and $y_{i}^{u} \in \mathcal{Y}^u$ is the corresponding label. For each class $c \in \mathcal{C} = \mathcal{C}^{s} \cup \mathcal{C}^{u}$, we have a class semantic vector $\boldsymbol{z}^{c}=\left[z_{1}^{c}, \ldots, z_{A}^{c}\right]^{\top}$ in which each item is annotated by human-beings according to the shared $A$ attributes. Only unseen classes $\mathcal{Y}^u$ are predicted for test images in CZSL, while both seen and unseen classes $ \mathcal{Y}=\mathcal{Y}^{s} \cup \mathcal{Y}^{u}$ should be recognized in GZSL.

\subsection{Dual Attention Block}
\noindent{\bf Multi-level Spatial Attention.}
In order to preserve the fine-grained features that may be lost in the high-level visual feature, we propose enriching the visual representation by explicitly fusing multi-level features. We project the visual features to the same scale as the features extracted from the second last block of the backbone for a visual-semantic trade-off. Additionally, a residual connection is utilized at this level to stabilize the training process. The final fused visual feature $\boldsymbol{v}_i$ is obtained by pixel-wise addition of the four parts, which will then be used to interact with attribute prototypes on spatial and channel dimension.

Building upon the attribute localization adopted by previous methods \cite{Huynh2020FineGrainedGZ,wang2021dual}, we propose to enhance the spatial relationship between visual features and semantic attributes through the learning of a region-attribute attention map. The trainable prototypes are shared across images with the random initialization, which will be used for attribute prediction. Denoting $\boldsymbol{p}_a$ as the attribute prototype, the attention map $\omega$ is calculated as,
\begin{equation}
\omega(\boldsymbol{p}_a, \boldsymbol{v}_i^r) = \frac{\exp \left(\boldsymbol{p}_{a}^{\top} \boldsymbol{W}_{1} \boldsymbol{v}_{i}^{r}\right)}{\sum_{r'=1}^{R} \exp \left(\boldsymbol{p}_{a}^{\top} \boldsymbol{W}_{1} \boldsymbol{v}_{i}^{r'}\right)},
\end{equation}
where $\boldsymbol{v}_i^r$ denotes the visual feature of the region $r$, $R$ is the number of feature grids, and $\boldsymbol{W}_1$ is a learnable matrix to measure the compatibility between each attribute prototype and region of the visual feature. We use a softmax activation function to normalize the attention weights $\{\omega(\boldsymbol{p}_a, \boldsymbol{v}_i^r)\}_{r=1}^R$ and drive the network to focus on the most relevant image regions for each attribute. The attention weights are utilized to compute the attribute feature for each image and thus achieve visual to semantic embedding, formulated as,
\begin{align}
\boldsymbol{k}_i^a = \sum_{r=1}^{R} \omega(\boldsymbol{p}_a, \boldsymbol{v}_i^r) \boldsymbol{v}_i^r,
\end{align}
where $\boldsymbol{k}_i^a$ represents the semantic feature of image $i$ that is relevant to the $a$-th attribute according to the prototype $\boldsymbol{p}_a$. 

\begin{figure}[t]
\begin{center}
\includegraphics[trim={80pt 40pt 80pt 35pt},clip,width=1\linewidth]{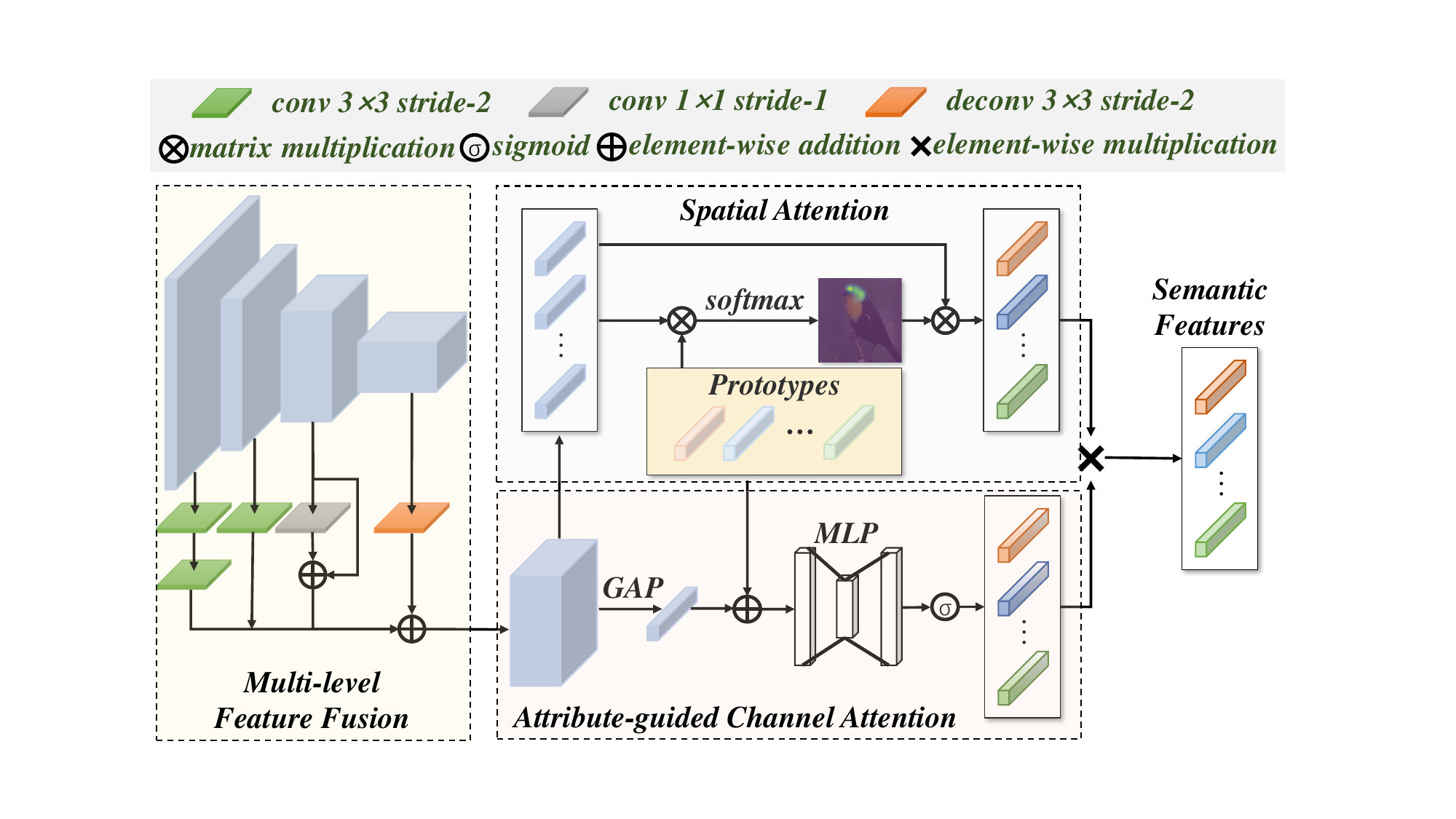}
\end{center}
\caption{The DAB models the visual-semantic relationship via Multi-level Spatial Attention and Attribute-guided Channel Attention. The disentangled semantic features are more beneficial for classification and knowledge transfer.}
\label{fig:DAB}
\end{figure}

\noindent{\bf Attribute-guided Channel Attention.}
The semantic features $\boldsymbol{k}_i^a$ extracted from spatial attention can become entangled when different attributes share similar attention areas. To overcome this shortcoming, we propose a novel channel attention mechanism that differentiates each semantic feature using attribute prototype guidance. As shown in Fig.\ref{fig:DAB}, we first leverage global average pooling on the visual feature $\boldsymbol{v}_i$ and normalize it to a normal distribution. Then we add the global visual feature with normalized attribute prototypes $\boldsymbol{p}_a$ to obtain a channel descriptor $\boldsymbol{q}_i^a$ of image $i$ for each attribute $a$ referred to \cite{hu2018squeeze}, formulated as,
\begin{align}
\boldsymbol{q}_i^a = norm(\boldsymbol{p}_a) + norm(\frac{1}{R} \sum_{r=1}^{R} \boldsymbol{v}_i^r).
\end{align}

To make use of the channel descriptor, we apply a bottleneck stacked by two fully-connected layers to capture channel-wise dependencies for each attribute. Additionally, we employ a gating mechanism with a sigmoid activation, and obtain the channel-wise attention weight $\boldsymbol{\eta}_i^a$ as follows, 
\begin{align}
\boldsymbol{\eta}_i^a = \boldsymbol{\sigma}(\textrm{\em MLP}(\boldsymbol{q}_i^a)) = \boldsymbol{\sigma}(\boldsymbol{W}_{3} \boldsymbol{\delta}(\boldsymbol{W}_{2} \boldsymbol{q}_i^a)),
\end{align}
where $\boldsymbol{\delta}$ refers to the ReLU function and $\boldsymbol{\sigma}$ denotes the sigmoid function. $\boldsymbol{W}_{2} \in \mathbb{R}^{\frac{C}{m} \times C}$ and $\boldsymbol{W}_{3} \in \mathbb{R}^{C \times \frac{C}{m}}$ are the bottleneck weights with a dimension reduction ratio $m$. The attention weight $\boldsymbol{\eta}_i^a$ can be considered as the soft-attention weight of channels associated with the $a$-th attribute. The disentangled semantic feature $\boldsymbol{h}_i^a$ is obtained by taking the element-wise multiplication of the semantic features and the channel weights as $\boldsymbol{h}_i^a = \boldsymbol{k}_i^a \cdot \boldsymbol{\eta}_i^a$.

\subsection{Semantic Interaction Transformer}
For zero-shot learning, transferring knowledge among attributes assumes significant importance owing to the scarcity of attributes. An effective way for knowledge transfer is to learn the relationship between attributes. Accordingly, we propose to model attribute relationship through a Semantic Interaction Transformer (SIT), which leverages an attention mechanism on the semantic features to obtain more robust attribute representations.

Let $\boldsymbol{H} \in \mathbb{R}^{B \times A \times C}$ denote a batch of semantic features, where B is the batch size, A is the number of attributes, and C is the dimension of semantic features. We first transform the dimension of $\boldsymbol{H}$ to $(B\times A, C)$ via reshape operator. Then $\boldsymbol{H}$ is feed as a sequence into a transformer encoder layer to learn the relationship among attributes. The operations can be formulated by,
\begin{align}
\boldsymbol{H}' &= \textrm{\em LN}(\textrm{\em MHSA}(\boldsymbol{H}) + \boldsymbol{H}), \\
\hat{\boldsymbol{H}} &= \textrm{\em LN}(\textrm{\em MLP}(\boldsymbol{H}') + \boldsymbol{H}').  
\end{align}

Since we do not always have a mini-batch of images during testing, SIT is employed as a plug-and-play module that can be removed during testing~\cite{hou2022batchformer,hou2022batchformerv2}. Meanwhile, we utilize a shared classifier before and after SIT to mitigate the gap in semantic features between training and testing.

\noindent{\bf Hyperspherical Classifier.}
To obtain the confidence of image $i$ having the attribute $a$, we calculate the similarity between semantic features $\boldsymbol{h}_i^a$ and prototypes $\boldsymbol{p}_a$ as,
\begin{equation}
e_i^a = \boldsymbol{p}_a^{\top} \boldsymbol{W}_4 \boldsymbol{h}_i^a,
\end{equation}
where $\boldsymbol{W}_4$ is an embedding matrix to project prototypes to the same dimension as semantic features.

Instead of utilizing dot product to compute the class logits $\boldsymbol{o}_i=\{o_i^c\}_{c=1}^C$ for image $i$, we project attribute scores $\boldsymbol{e}_i = \{e_i^a\}_{a=1}^A$ and class semantics $\boldsymbol{z}^c$ onto the hyperspherical space via $l_2$-normalization and employ the scaled cosine similarity to achieve better generalization ability \cite{skorokhodov2020class},
\begin{equation}
o_i^c = \left(\gamma \cdot \frac{\boldsymbol{z}^c}{\|\boldsymbol{z}^c\|}\right)^\top \left(\gamma \cdot \frac{\boldsymbol{e}_i}{ \|\boldsymbol{e}_i\|}\right),
\end{equation}
where $\gamma$ is a hyperparameter to control the range of scale.

\subsection{Global Classification Branch}
Noting the potential for latent features to supplement human-defined attributes \cite{peng2017joint,jiang2017learning,Liu2019AttributeAF}, we adopt a simple yet effective approach that leverages a global classification branch without constraints on class semantics during training. First, we obtain the visual feature from the last block of the backbone and apply global average pooling to generate a global feature. Next, we use a linear layer to project the global feature to a class logits $g_i$. The predicted class logits $g_i$ is then used to compute cross-entropy with the corresponding class $y_i$. After obtaining the class logits, we combine the results with attribute-based classification to improve performance on seen classes in GZSL during testing. Additionally, we find that jointly learning latent features and human-defined attributes can improve semantic attribute prediction.

\subsection{Loss Function}
We employ the attribute-based classification loss with self-calibration term and global classification loss to optimize the parameters in our DRMN.

\noindent{\bf Attribute-based Classification Loss.}
Since the classification results of ZSL is highly related to the attribute prediction, we utilize attribute-based classification loss $\mathcal{L}_{\textrm{\em AC}}$ with self-calibration term \cite{Huynh2020FineGrainedGZ} to project the attribute score of an image near its corresponding class semantic vector and avoid over-fitting on seen classes, which reads,
\begin{equation}
\begin{split}
\mathcal{L}_{\textrm{\em AC}}&=-\frac{1}{N} \sum_{i=1}^{N} \sum_{c \in \mathcal{C}^s} \mathbb {I}_{\left[c=y_i^s\right]} \log \frac{\exp \left(o_i^c \right)}{\sum_{\hat{c} \in \mathcal{C}^s} \exp \left(o_i^{\hat{c}} \right)}\\
-\lambda_{\textrm{\em SC}} &\frac{1}{N} \sum_{i=1}^{N} \sum_{c^{\prime} \in \mathcal{C}^u} \log \frac{\exp \left(o_i^{c^{\prime}} + \mathbb {I}_{\left[c^{\prime}\in\mathcal{C}^u\right]}\right)}{\sum_{\hat{c} \in \mathcal{C}} \exp \left(o_i^{\hat{c}} + \mathbb {I}_{\left[\hat{c}\in\mathcal{C}^u\right]}\right)},
\end{split}
\end{equation}
where $N$ is the batch size, $o_i^c$ is the class logit and $y_i^s$ is the ground-truth label for image $i$. $\mathbb {I}_{[\cdot]}$ is the indicator function.

\noindent{\bf Global Classification Loss.}
A global classification loss is utilized to align the global feature and class label. Given a batch of $N$ training images with their corresponding class label $y_i^s$, we denote the prediction as $g_i^c$ and define $\mathcal{L}_{\textrm{\em GC}}$ as,
\begin{equation}
\mathcal{L}_{\textrm{\em GC}}=-\frac{1}{N} \sum_{i=1}^{N} \sum_{c=1}^{C} \mathbb {I}_{\left[c=y_i^s\right]} \log \frac{\exp \left(g_i^{c} \right)}{\sum_{\hat{c} \in \mathcal{C}} \exp \left(g_i^{\hat{c}}\right)}.
\end{equation}

Finally, the overall loss function of our method reads,
\begin{equation}
\mathcal{L}_{\textrm{\em total}}=\mathcal{L}_{\textrm{\em AC}}+\lambda_{\textrm{\em GC}} \mathcal{L}_{\textrm{\em GC}},
\end{equation}
where $\lambda_{\textrm{\em GC}}$ is the hyperparameter to control its corresponding loss term.

\subsection{Zero-Shot Prediction}
During inference, we determine the predicted label for CZSL simply by selecting the class with the maximum score. For GZSL, we use an ensemble strategy to combine the results of global classification and attribute-based classification for the final prediction. Since the two branches treat the problem from different perspectives, the output results are complementary to each other. The details of the ensemble strategy can be found in \ref{alg:Algorithm1}.

\begin{algorithm}[h]
\caption{Ensemble Strategy}
\label{alg:Algorithm1}
\hspace*{0.02in} {\bf Input:}\\
\hspace*{0.2in} $\boldsymbol{o}_i: \{o_i^c\}_{c=1}^C$ the output of attribute-based classification \\
\hspace*{0.2in} $\boldsymbol{g}_i: \{g_i^c\}_{c=1}^{C}$ the output of global classification\\
\hspace*{0.2in} $\beta$ : the weight of ensemble\\
\hspace*{0.02in} {\bf Output:} \\
\hspace*{0.2in} the predicted label $y$
\begin{algorithmic}

\STATE $y_1 \leftarrow \arg \max (\boldsymbol{o}_i + \mathbb{I}_{C_u}(\boldsymbol{o}_i))$
\IF{$y_1 \in C^s$}
\STATE $\hat{\boldsymbol{o}}_i \leftarrow{\textrm{\em softmax}(\boldsymbol{o}_i)}$
\STATE $\hat{\boldsymbol{g}}_i \leftarrow{\textrm{\em softmax}(\boldsymbol{g}_i)}$
\STATE $y \leftarrow {\arg \max (\beta \times \hat{\boldsymbol{o}}_i + (1-\beta) \times \hat{\boldsymbol{g}}_i)}$
\ELSE 
\STATE $y \leftarrow{y_1}$
\ENDIF 
\RETURN $y$
\end{algorithmic}
\end{algorithm}

\begin{table*}[htb!]
\small
\begin{center}
\resizebox{0.85\linewidth}{!}{
  \begin{tabular}{c|cccc|cccc|cccc}
  \hline
  \multirow{3}{*}{Methods}&\multicolumn{4}{c|}{CUB}&\multicolumn{4}{c|}{SUN}&\multicolumn{4}{c}{AwA2}\\
  &CZSL&\multicolumn{3}{c|}{GZSL}&CZSL&\multicolumn{3}{c|}{GZSL}&CZSL&\multicolumn{3}{c}{GZSL}\\
  &$Acc$&$U$&$S$&$H$&$Acc$&$U$&$S$&$H$&$Acc$&$U$&$S$&$H$\\
  \hline
  \multicolumn{13}{c}{generative-based methods}\\
  \hline
  f-CLSWGAN \cite{Xian2018FeatureGN} &57.3&43.7&57.7&49.7&60.8&42.6&36.6&39.4&68.2&57.9&61.4&59.6\\
  f-VAEGAN-D2 \cite{xian2019f} &61.0&48.4&60.1&53.6&64.7&45.1&38.0&41.3&71.1&57.6&70.6&63.5\\
  OCD-CVAE \cite{keshari2020generalized} &60.3&44.8&59.9&51.3&63.5&44.8&\textbf{\color{red}42.9}&\textbf{\color{blue}43.8}&71.3&59.5&73.4&65.7\\
  LsrGAN \cite{Vyas2020LeveragingSA} &60.3&48.1&59.1&53.0&62.5&44.8&37.7&40.9&-&-&-&-\\
  E-PGN \cite{Yu2020EpisodeBasedPG} &72.4&52.0&61.1&56.2&-&-&-&-&73.4&52.6&83.5&64.6\\
  CE-GZSL \cite{Han2021ContrastiveEF} &77.5&63.9&66.8&65.3&63.3&48.8&38.6&43.1&70.4&63.1&78.6&70.0\\
  FREE \cite{Chen2021FREE} &-&55.7&59.9&57.7&-&47.4&37.2&41.7&-&60.4&75.4&67.1\\
  HSVA \cite{Chen2021HSVA} &-&52.7&58.3&55.3&-&48.6&39.0&43.3&-&56.7&79.8&66.3\\
  ICCE \cite{kong2022compactness} &-&67.3&65.5&66.4&-&-&-&-&-&65.3&82.3&\textbf{\color{blue}72.8}\\
  \hline
  \multicolumn{13}{c}{embedding-based methods}\\
  \hline
  AREN \cite{xie2019attentive} &71.8&63.2&69.0&66.0&60.6&40.3&32.3&35.9&67.9&54.7&79.1&64.7\\
  LFGAA \cite{Liu2019AttributeAF}&67.6&36.2&\textbf{\color{red}80.9}&50.0&61.5&18.5&\textbf{\color{blue}40}&25.3&68.1&27.0&\textbf{\color{red}93.4}&41.9\\
  DAZLE \cite{Huynh2020FineGrainedGZ} &66.0&56.7&59.6&58.1&59.4&52.3&24.3&33.2&67.9&60.3&75.7&67.1\\
  RGEN \cite{xie2020region} &76.1&60.0&73.5&66.1&63.8&44&31.7&36.8&\textbf{\color{blue}73.6}&\textbf{\color{red}67.1}&76.5&71.5\\
  APN \cite{Xu2020AttributePN} &72.0&65.3&69.3&67.2&61.6&41.9&34.0&37.6&68.4&56.5&78.0&65.5\\
  GEM \cite{Liu2021GoalOrientedGE} &\textbf{\color{blue}77.8}&64.8&77.1&70.4&62.8&38.1&35.7&36.9&67.3&64.8&77.5&70.6\\
  DPPN \cite{wang2021dual} &-&\textbf{\color{blue}70.2}&77.1&\textbf{\color{blue}73.5}&-&47.9&35.8&41.0&-&63.1&\textbf{\color{blue}86.8}&\textbf{\color{red}73.1}\\
  TransZero \cite{Chen2022TransZeroAT} &76.8&69.3&68.3&68.8&65.6&\textbf{\color{blue}52.6}&33.4&40.8&70.1&61.3&82.3&70.2\\
  MSDN \cite{chen2022msdn} &76.1&68.7&67.5&68.1&\textbf{\color{blue}65.8}&52.2&34.2&41.3&70.1&62.0&74.5&67.7\\ 
  DRMN(Ours) &\textbf{\color{red}82.5}&\textbf{\color{red}75.5}&\textbf{\color{blue}78.1}&\textbf{\color{red}76.8}&\textbf{\color{red}66.9}&\textbf{\color{red}54.8}&39.3&\textbf{\color{red}45.8}&\textbf{\color{red}74.6}&\textbf{\color{blue}66.1}&77.8&71.5\\
  \hline
  \end{tabular}
  }
\end{center}
  \caption{Results (\%) of the state-of-the-art CZSL and GZSL methods on CUB, SUN, and AwA2. $Acc$ is the top-1 accuracy for CZSL. $U$ and $S$ is the top-1 accuracy for unseen and seen classes, respectively. $H$ is the harmonic mean calculated from $U$ and $S$ for an accuracy trade-off. The first part is generative-based methods, and the second part is embedding-based methods. The best and second-best results are marked in \textbf{\color{red}Red} and \textbf{\color{blue}Blue}, respectively.}
  \label{tab:compare-with-sota}
\end{table*}

%% file: sec/4_experiments.tex
\section{Experiments} 
\label{sec:experiments}
\noindent{\bf Datasets.}
We evaluate our method on three widely used ZSL benchmarks, in which CUB \cite{Welinder2010CaltechUCSDB2} and SUN \cite{Patterson2012SUNAD} are two fine-grained datasets while AwA2 \cite{Xian2017ZeroShotLC} is a coarse-grained one. We use the Proposed Split (PS) for training and testing following \cite{Xian2017ZeroShotLC}. Specifically, CUB is annotated with 312 attributes and contains 11,788 images from 200 types of birds, in which 150 species are selected as seen classes and the rest belongs to unseen classes. SUN is a scene dataset and has 14,340 images with 102 attributes. It involves 645 seen classes and 72 unseen classes. AwA2 only has 85 attributes, including 37,322 images from 50 kinds of animals with 40 seen classes and 10 unseen classes. 

\noindent{\bf Evaluation Protocols.}
We use average per-class top-1 accuracy following \cite{Xian2017ZeroShotLC} in both conventional zero-shot learning (CZSL) and generalized zero-shot learning (GZSL). For CZSL setting, only unseen classes are used for evaluation and we denote the top-1 accuracy as $Acc$. In GZSL setting, we calculate the accuracy for both seen and unseen samples during inference which is denoted as $S$ and $U$, respectively. For an accuracy trade-off between seen and unseen performance, the harmonic mean (denoted as $H$) is computed as $H = (2 \times S \times U) / (S + U).$

\noindent{\bf Implementation Details.}
We use a ResNet101 \cite{he2016deep} pre-trained on ImageNet as the backbone to extract visual features from each image. The dimension reduction ratio $m$ is set as 4. Adam optimizer \cite{Adam} is utilized to train the model. The learning rate is set as 0.001, which is decayed every 10 epochs with a decay factor 0.5. The loss weights $\{\lambda_\textrm{\em SC}, \lambda_\textrm{\em GC}\}$ are set to be $\{0.1, 0.6\}$ for CUB, $\{0.1, 0.6\}$ for SUN, and $\{0.2, 0.5\}$ for AwA2, respectively. For the ensemble weight $\beta$, we set it as 0.3 for CUB, 0.4 for SUN, and 0.3 for AwA2 according to our hyperparameter analysis. 

\subsection{Comparison with State-of-the-Arts}
\noindent{\bf Conventional Zero-Shot Learning.}
We first conduct a comparison of our DRMN with the state-of-the-art methods in CZSL setting. As presented in Table~\ref{tab:compare-with-sota}, DRMN achieves the highest accuracies of $82.5\%$, $66.9\%$ and $74.6\%$ on CUB, SUN, and AwA2, respectively. These results signify that the dual relation (\textit{i.e.}, visual-semantic relationship and semantic-semantic relationship) learned by our method has the potential to attain remarkable generalization ability for unseen classes. Moreover, the outcomes also indicate that the more effective visual-semantic interaction and generalized semantic representation play a crucial role in facilitating knowledge transfer.

\begin{figure*}[t]
\begin{center}
\includegraphics[width=1\linewidth]{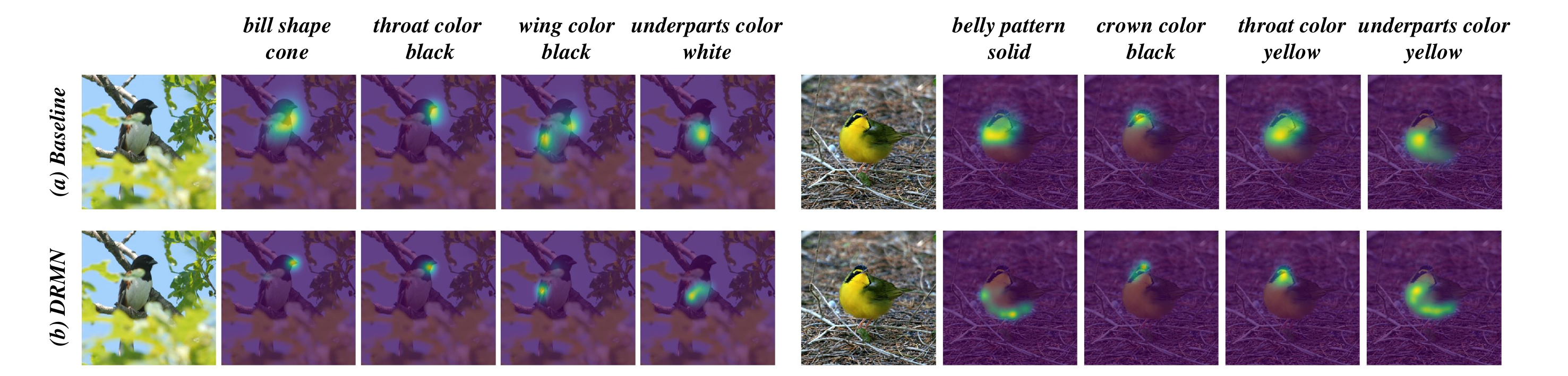}
\end{center}
\vspace{-2mm}
\caption{Visualization of attention maps for the baseline and our DRMN.}
\label{fig:attention_map}
\end{figure*}

\noindent{\bf Generalized Zero-Shot Learning.}
We also compare our DRMN with the state-of-the-art GZSL methods in Table~\ref{tab:compare-with-sota}. One can check that numerous previous methods encounter difficulties in identifying unseen classes, particularly for CUB and SUN. However, our DRMN achieves the best generalization ability to unseen classes on CUB and SUN, with a comparable classification accuracy to previous methods on AwA2. This has resulted in new state-of-the-art harmonic mean $H$ of $76.8\%$ and $45.8\%$ on CUB, SUN, respectively, as well as a comparable H of $71.5\%$ on AwA2. We analyse that the benefits of DRMN stem from the more effective visual-semantic interaction within the DAB, which enhances the discriminability of semantic features. Furthermore, the SIT learns a more comprehensive semantic representation that is beneficial for knowledge transfer and attribute prediction. Additionally, the global classification branch complements the human-defined attributes, leading to further performance improvements. The aforementioned advantages collectively demonstrate the superiority and enormous potential of exploring intricate relationship between visual and semantic features for ZSL.

\begin{table}[t]
\begin{center}
\resizebox{0.8\linewidth}{!}{
\begin{tabular}{cc|cc|cc}
\hline
\multicolumn{2}{c|}{Method}&\multicolumn{2}{c|}{CUB}&\multicolumn{2}{c}{SUN}\\
DAB&SIT&$Acc$&$H$&$Acc$&$H$\\
\hline
\multicolumn{2}{c|}{baseline}&71.5 &64.9 &60.6 &34.8\\
\hline
 \checkmark & &81 &74.5 &64.4 &39.2\\
 &\checkmark &74.9 &67.5 &62.9 &35.9\\
 \checkmark &\checkmark &81.2 &75.3 &64.5 &40\\
 \hline
 \multicolumn{2}{c|}{+ global branch}&82.5 &76.1 &66.9 &42.3\\
\multicolumn{2}{c|}{+ ensemble strategy}&82.5 &76.8 &66.9 &45.8\\
\hline
\end{tabular}
}
\end{center}
\vspace{-1.5mm}
\caption{Ablation studies for different modules (\textit{i.e.}, Dual Attention Block (DAB), Semantic Interaction Transformer (SIT), global classification branch and ensemble strategy) of our DRMN. The baseline only conducts spatial attention for visual to semantic embedding and utilize the hyperspherical classifier for classification.}
\label{tab:ablation1}
\end{table}

\begin{table}[t]
\begin{center}
\resizebox{0.7\linewidth}{!}{
\begin{tabular}{cc|cc|cc}
\hline
\multicolumn{2}{c|}{Method}&\multicolumn{2}{c|}{CUB}&\multicolumn{2}{c}{SUN}\\
MFF&ACA&$Acc$&$H$&$Acc$&$H$\\
\hline
\multicolumn{2}{c|}{baseline}&71.5 &64.9 &60.6 &34.8\\
\hline
 \checkmark & &79.7 &73.3 &63.5 &38.5\\
 &\checkmark &74 &66.9 &61.9 &39\\
 \checkmark &\checkmark &81 &74.5 &64.4 &39.2\\
\hline
\end{tabular}
}
\end{center}
\vspace{-1.5mm}
\caption{Ablation studies for different modules (\textit{i.e.}, Multi-level Feature Fusion (MFF) and Attribute-guided Channel Attention (ACA)) of our DAB.}
\label{tab:ablation2}
\end{table}

\subsection{Ablation Study}
To evaluate the effectiveness of the proposed DRMN, we conduct ablation studies to analyze the influence of different modules and operations, as shown in Table~\ref{tab:ablation1}. One can see that the Dual Attention Block (DAB) significantly improves $Acc$/$H$ by $9.5\%$/$9.6\%$ and $3.8\%$/$4.4\%$ on CUB and SUN, respectively. These gains demonstrate that the DAB conducts a more effective visual-semantic interaction and decouples semantic features for more precise attribute prediction. For semantic relationship modeling, we evaluate the effectiveness of the Semantic Interaction Transformer (SIT), which can improve $Acc/H$ over the baseline by $3.4\%$/$2.6\%$ and $2.3\%$/$1.1\%$ on CUB and SUN, respectively. Additionally, it can enhance the performance of $H$ beyond the addition of the DAB. These results reveal that a more generalized semantic representation is beneficial for knowledge transfer. Regarding the global classification branch, we observe $Acc/H$ improvements of $1.3\%$/$0.8\%$ and $2.4\%$/$2.3\%$ on CUB and SUN, respectively, indicating that learning latent features jointly with human-defined attributes can promote attribute prediction. Furthermore, by applying the ensemble operation, we achieve additional gains of $0.7\%$ and $3.5\%$ on $H$. 

To further analyze the impact of the Dual Attention Block, we conduct ablation studies to evaluate the influence of Multi-level Feature Fusion (MFF) and Attribute-guided Channel Attention (ACA), as shown in Table~\ref{tab:ablation2}. Our results demonstrate that the MFF can bring a significant $Acc$/$H$ gains of $8.2\%$/$8.4\%$ on CUB and $2.9\%$/$3.7\%$ on SUN over the baseline. Similarly, the ACA can also boost performance beyond the baseline by a large margin. These findings suggest that incorporating multi-level features and decoupling semantic representations can enable more effective visual-semantic interaction.

\begin{figure}[t]
\begin{center}
\includegraphics[trim={110pt 110pt 135pt 150pt},clip,width=0.9\linewidth]{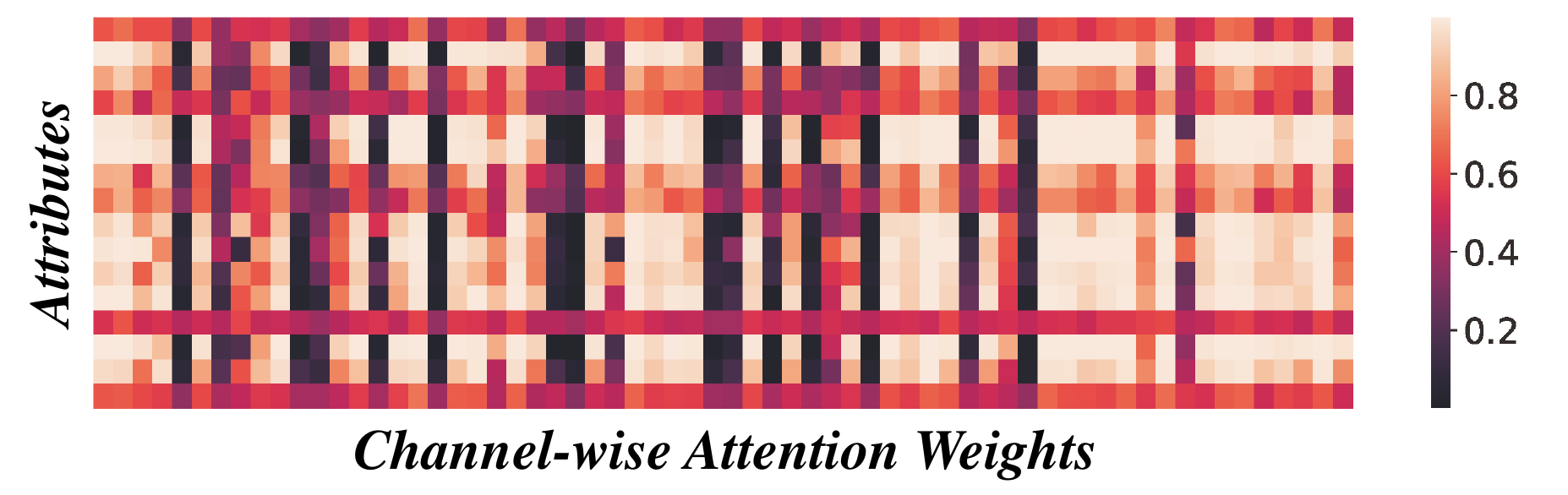}
\end{center}
\vspace{-2mm}
\caption{Visualization of channel-wise attention weights learned by our Attribute-guided Channel Attention. The channel-wise attention weights and attributes are randomly selected.}
\label{fig:channel_att}
\end{figure}
\vspace{-1mm}

\begin{figure*}[t]
  \centering
  \begin{subfigure}{0.48\linewidth}
    \includegraphics[trim={50pt 0pt 50pt 80pt},clip,width=1\linewidth]{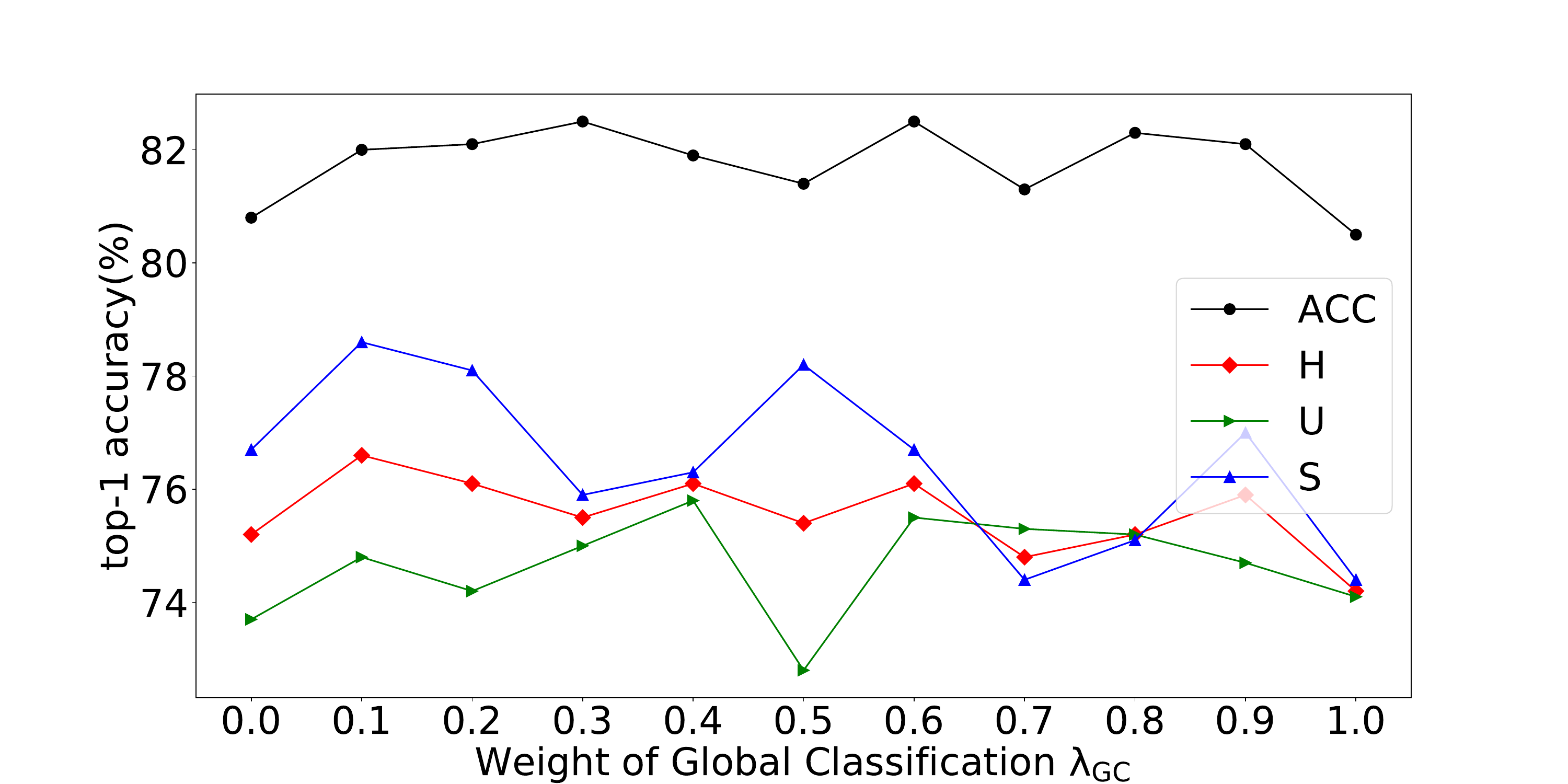}
    \caption{CUB}
  \end{subfigure}
  \hfill
  \begin{subfigure}{0.48\linewidth}
    \includegraphics[trim={50pt 0pt 50pt 80pt},clip,width=1\linewidth]{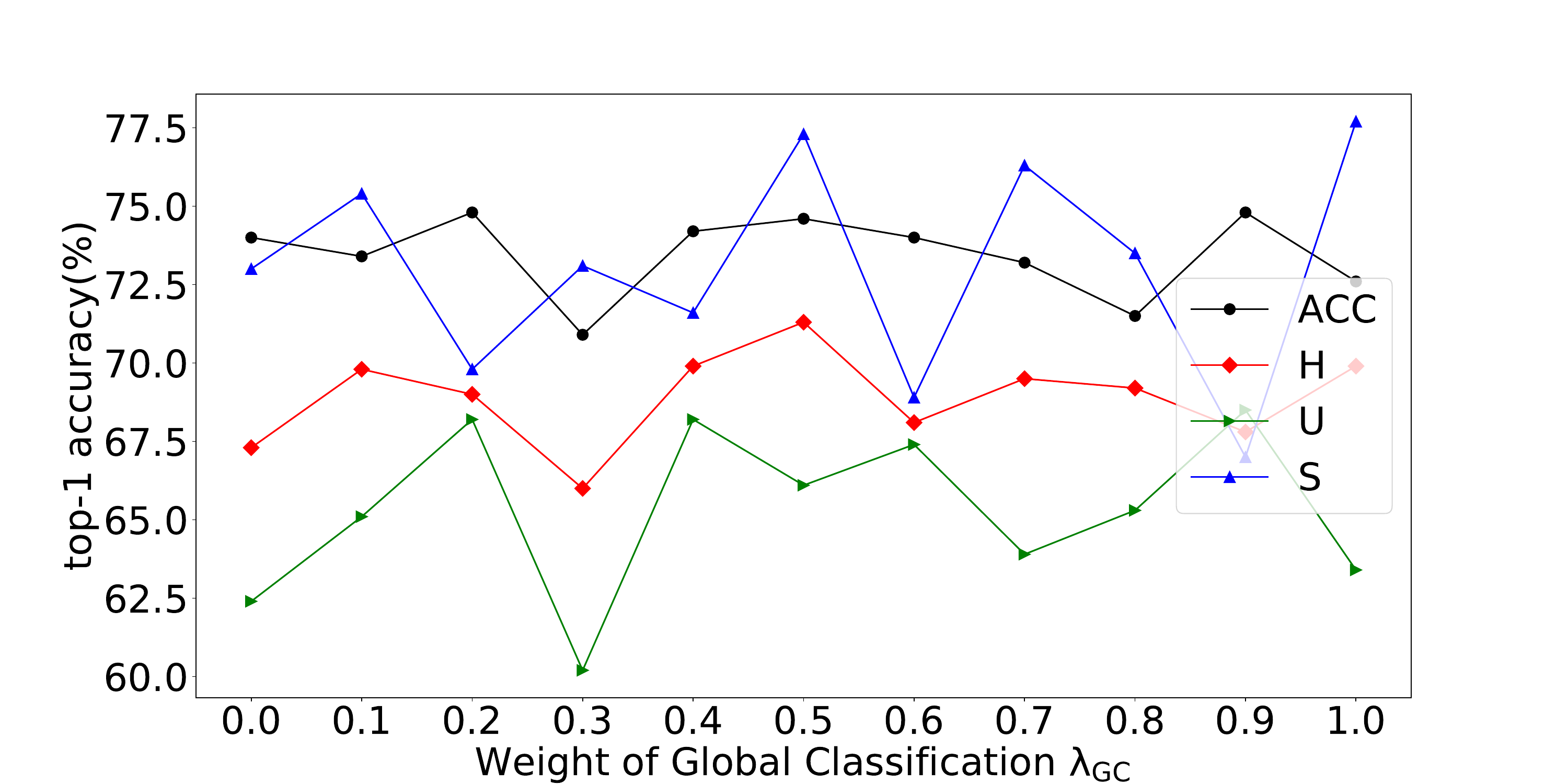}
    \caption{AwA2}
  \end{subfigure}
  \vspace{-3mm}
  \caption{The effects of global classification branch weight $\lambda_{\textrm{\em GC}}$.}
\label{fig:global_branch}
\end{figure*}

\begin{figure*}[t]
  \centering
  \begin{subfigure}{0.48\linewidth}
    \includegraphics[trim={50pt 0pt 50pt 80pt},clip,width=1\linewidth]{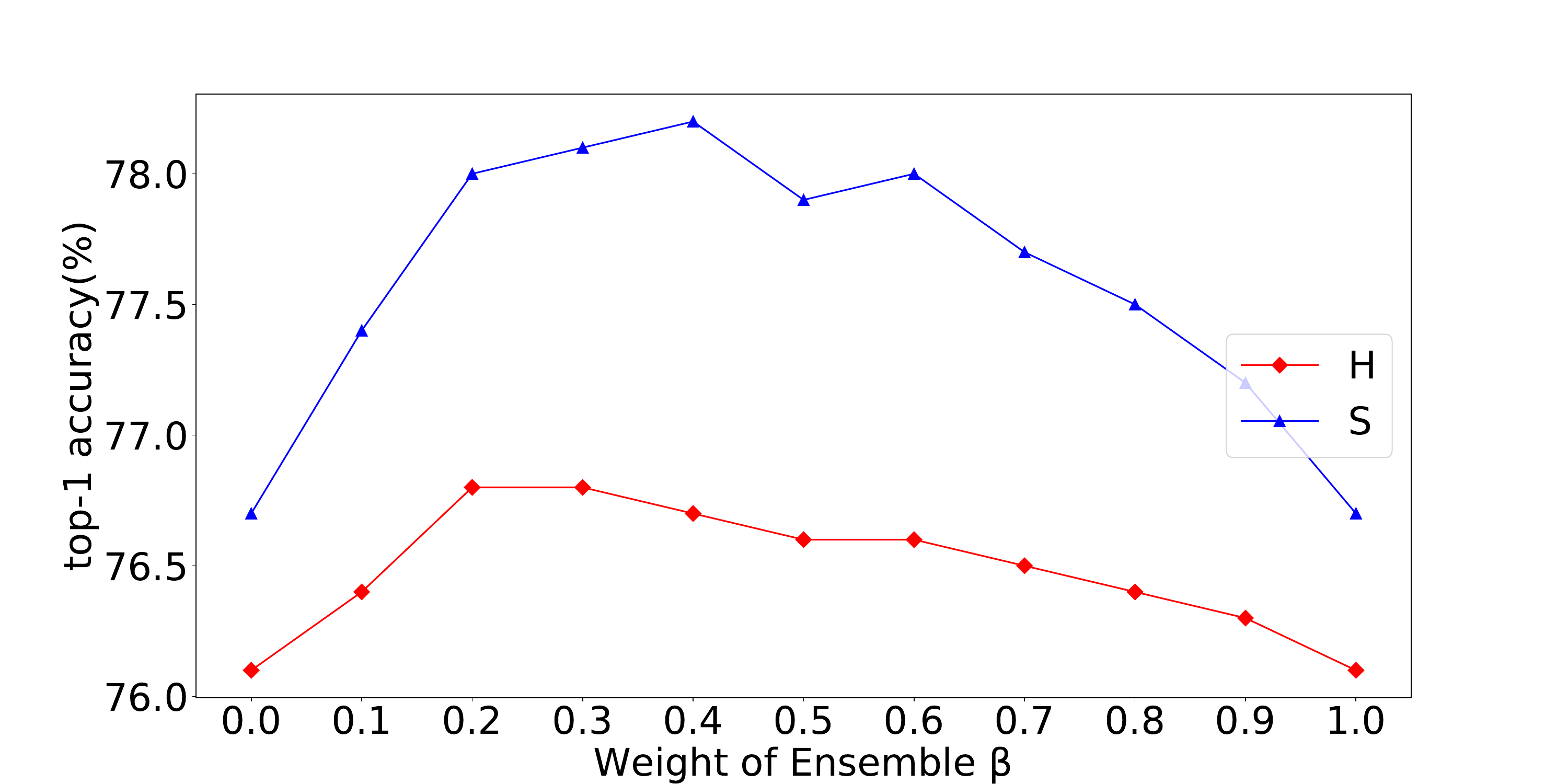}
    \caption{CUB}
  \end{subfigure}
  \hfill
  \begin{subfigure}{0.48\linewidth}
    \includegraphics[trim={50pt 0pt 50pt 80pt},clip,width=1\linewidth]{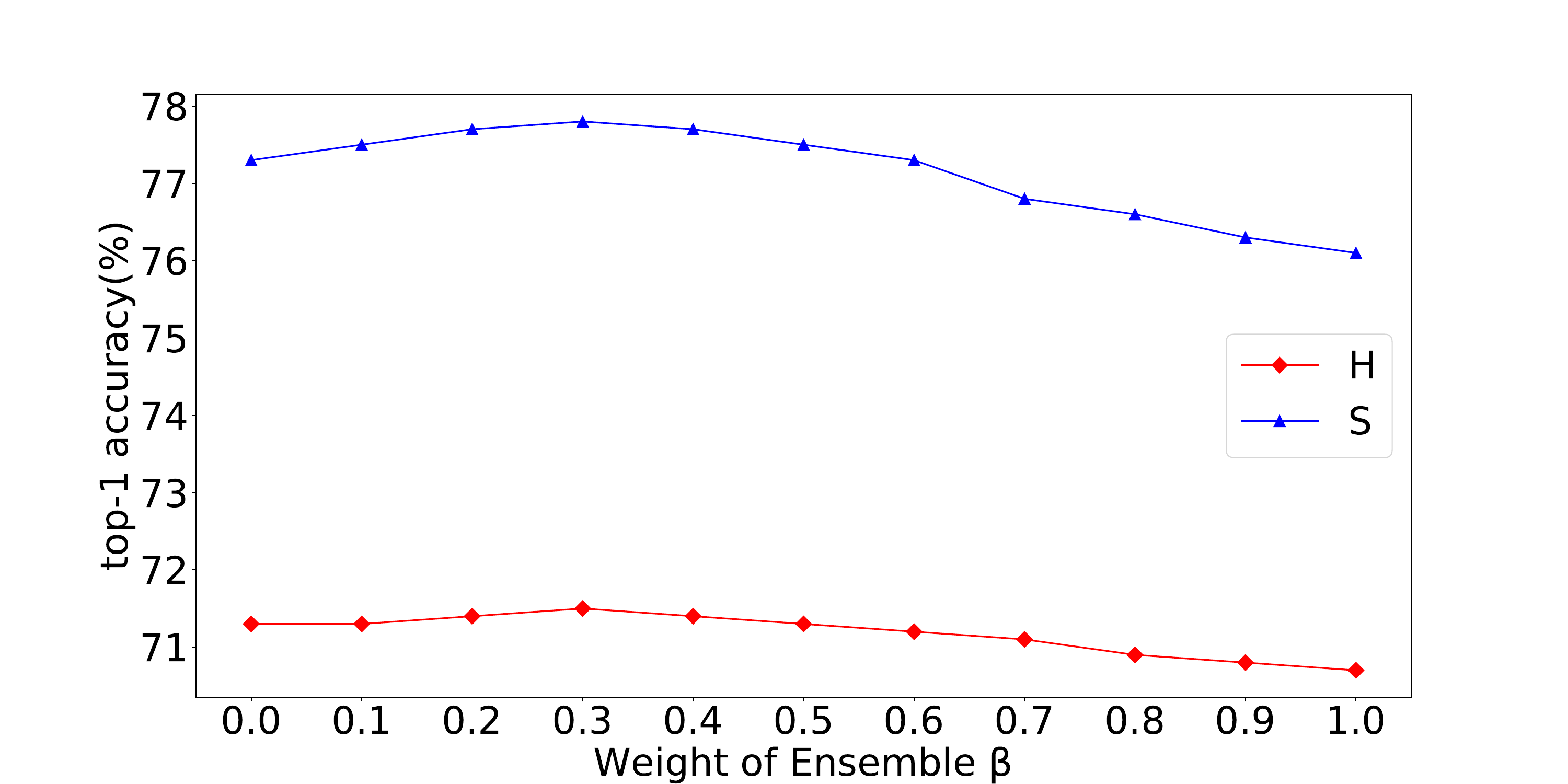}
    \caption{AwA2}
  \end{subfigure}
  \vspace{-3mm}
  \caption{The effects of ensemble weight $\beta$.}
\label{fig:ensemble}
\end{figure*}

\subsection{Qualitative Results}
\noindent{\bf Visualization of Attention Maps.}
To demonstrate the effectiveness of DRMN in attribute localization, we visualize the attention maps learned by both the baseline and our DRMN. As illustrated in Fig.\ref{fig:attention_map}, the misalignment between attributes and image regions sometimes occurs in the baseline while DRMN can localize attribute-related regions more accurately. This benefit mainly comes from the fusion of multi-level visual features, which facilitates fine-grained parts localization (e.g., bill shape cone and crown color black). Moreover, the decoupled and generalized semantic representation is helpful for attribute prediction.

\noindent{\bf Visualization of Channel Attention.}
To show the effectiveness of Attribute-Guided Channel Attention for decoupling semantic features, we visualize the channel-wise attention weights for different attributes. As shown in Fig.\ref{fig:channel_att}, we can see that different attributes have a different channel attention weights, which will help decouple the semantic features extracted from spatial attention. 

\vspace{-1mm}
\subsection{Hyperparameter Analysis}
\noindent{\bf Effects of Global Classification Loss.}
$\lambda_{\textrm{\em GC}}$ is utilized to weigh the importance of the global classification loss $\mathcal{L}_{\textrm{\em GC}}$. The global classification branch is trained with the attribute-based classification branch jointly, and thus it will have an influence on the final classification results. As shown in Fig.\ref{fig:global_branch}, one can see that the global classification loss helps learn a better attribute representation for classification, and a proper loss term can achieve better accuracy of unseen classes. Based on our observation in Fig. \ref{fig:global_branch}, we choose 0.6 for CUB and 0.5 for AwA2, respectively.

\noindent{\bf Effects of Ensemble Weight.}
$\beta$ is employed to weigh the importance of attribute-based classification results and global classification results on seen classes. We only utilize the attribute-based classification for the prediction of unseen classes because the global classification branch has no ability for knowledge transfer. As shown in Fig.\ref{fig:ensemble}, one can see that using the results from the global classification branch or the attribute-based classification branch individually is not the best choice for seen class classification. Since the attribute-based classification and global classification are complementary to each other, a proper weight $\beta$ will be a good trade-off of the two branches. Thus, the ensembled results can lead to better performance. Based on our experiments, we set $\beta$ as 0.3 for CUB and AwA2. 

\vspace{-1mm}

%% file: sec/5_conclusion.tex
\section{Conclusion}
\vspace{-1mm}
In this paper, we propose a Dual Relation Mining Network (DRMN) to jointly mine visual-semantic relationship and semantic-semantic relationship for ZSL. Specifically, our DRMN comprises a Dual Attention Block (DAB) for visual-semantic relationship mining and a Semantic Interaction Transformer (SIT) for semantic relationship modeling. Within the DAB, we enhance the visual feature via multi-level feature fusion and employ dual attention for visual to semantic embedding. To explore semantic relationship, we use SIT to enhance the generalization of attribute representations among images. We also introduce a global classification branch as a complement to human-defined semantic attributes. As a result, our DRMN achieves effective visual-semantic interaction and generalized semantic representation for knowledge transfer. Extensive experiments on three ZSL benchmarks show the effectiveness of our DRMN.